\documentclass[letterpaper, 10 pt, conference]{ieeeconf}  % Comment this line out if you need a4paper

\IEEEoverridecommandlockouts                              % This command is only needed if 
\usepackage[pdftex]{graphicx} % for arXiv

\usepackage{epsfig} % for postscript graphics files
\usepackage{amsmath} % assumes amsmath package installed
\usepackage{amssymb}  % assumes amsmath package installed
\usepackage{url}

\newcommand{\ie}{i.e., }
\newcommand{\eg}{e.g., }
\newcommand{\st}{\mathrm{s.t.} \;\;\;}
\newcommand{\argmin}{\mathop{\rm arg~min}\limits}
\newcommand{\SP}{\;\;}
\newcommand{\SPP}{\SP\SP}

\title{\LARGE \bf
Fast MILP-based Task and Motion Planning for Pick-and-Place\\
with Hard/Soft Constraints of Collision-Free Route
}

\author{Takuma Kogo, Kei Takaya, and Hiroyuki Oyama% <-this % stops a space
\thanks{The authors are with Data Science Research Laboratories, NEC Corporation, Japan.}%
}
\begin{document}

\maketitle
\thispagestyle{empty}
\pagestyle{empty}
\urlstyle{}
%\setlength{\abovedisplayskip}{3pt}
%\setlength{\belowdisplayskip}{3pt}

%%%%%%%%%%%%%%%%%%%%%%%%%%%%%%%%%%%%%%%%%%%%%%%%%%%%%%%%%%%%%%%%%%%%%%%%%%%%%%%%
\begin{abstract}
We present new models of optimization-based task and motion planning (TAMP) for robotic pick-and-place (P\&P), which plan action sequences and motion trajectory with low computational costs. We improved an existing state-of-the-art TAMP model integrated with the collision avoidance, which is formulated as a mixed-integer linear programing (MILP) problem. To enable the MILP solver to search for solutions efficiently, we introduced two approaches leveraging features of collision avoidance in robotic P\&P. The first approach reduces number of binary variables, which are related to the collision avoidance of delivery objects, by reformulating them as continuous variables with additional hard constraints. These hard constraints maintain consistency by conditionally propagating binary values, which are related to the carry action state and collision avoidance of robots, to the reformulated continuous variables. The second approach is more aware of the branch-and-bound method, which is the fundamental algorithm of modern MILP solvers. This approach guides the MILP solver to find integer solutions with shallower branching by adding a soft constraint, which softly restricts a robot’s routes around delivery objects. We demonstrate the effectiveness of the proposed approaches with a modern MILP solver.

\end{abstract}

%%%%%%%%%%%%%%%%%%%%%%%%%%%%%%%%%%%%%%%%%%%%%%%%%%%%%%%%%%%%%%%%%%%%%%%%%%%%%%%%
%%%%%%%%%%%%%%%%%%%%%%%%%%%%%%%%%%%%%%%%%%%%%%%%%%%%%%%%%%%%%%%%%%%%%%%%%%%%%%%%
%%%%%%%%%%%%%%%%%%%%%%%%%%%%%%%%%%%%%%%%%%%%%%%%%%%%%%%%%%%%%%%%%%%%%%%%%%%%%%%%

\section{Introduction}

%%%%%%%%%%%%%%%%%%%%%%%%%%%%%%%%%%%%%%%%%%%%%%%%%%%%%%%%%%%%%%%%%%%%%%%%%%%%%%%%

\subsection{Background}
Automation improves companies' productivity and competitiveness. Therefore, robots have been in increasing demand in industries, logistics, and other emerging domains for several years \cite{c01}. Currently, a major task of robots is handling (\eg pick-and-place, packing, palletizing, etc.) \cite{c01}. In particular, pick-and-place (P\&P), which is the sequential task of picking objects up from one location and placing them in a desired location, is a fundamental and essential task in many fields.

A basic approach to enable a robot to perform P\&P is {\it teaching} where a human expert programs the robot's behavior with primitive commands on a step-by-step basis. {\it Teaching} is applicable to a variety of workspaces. However, slight changes to P\&P specifications (\eg layout of workspace, delivery object, etc.) require update works of {\it teaching} which is much time consuming. Therefore, {\it teaching-less} approaches have been desired to mitigate the aforementioned difficulties.

\begin{figure}[!t]
    \centering
    \includegraphics[keepaspectratio,clip,width=0.83\hsize]{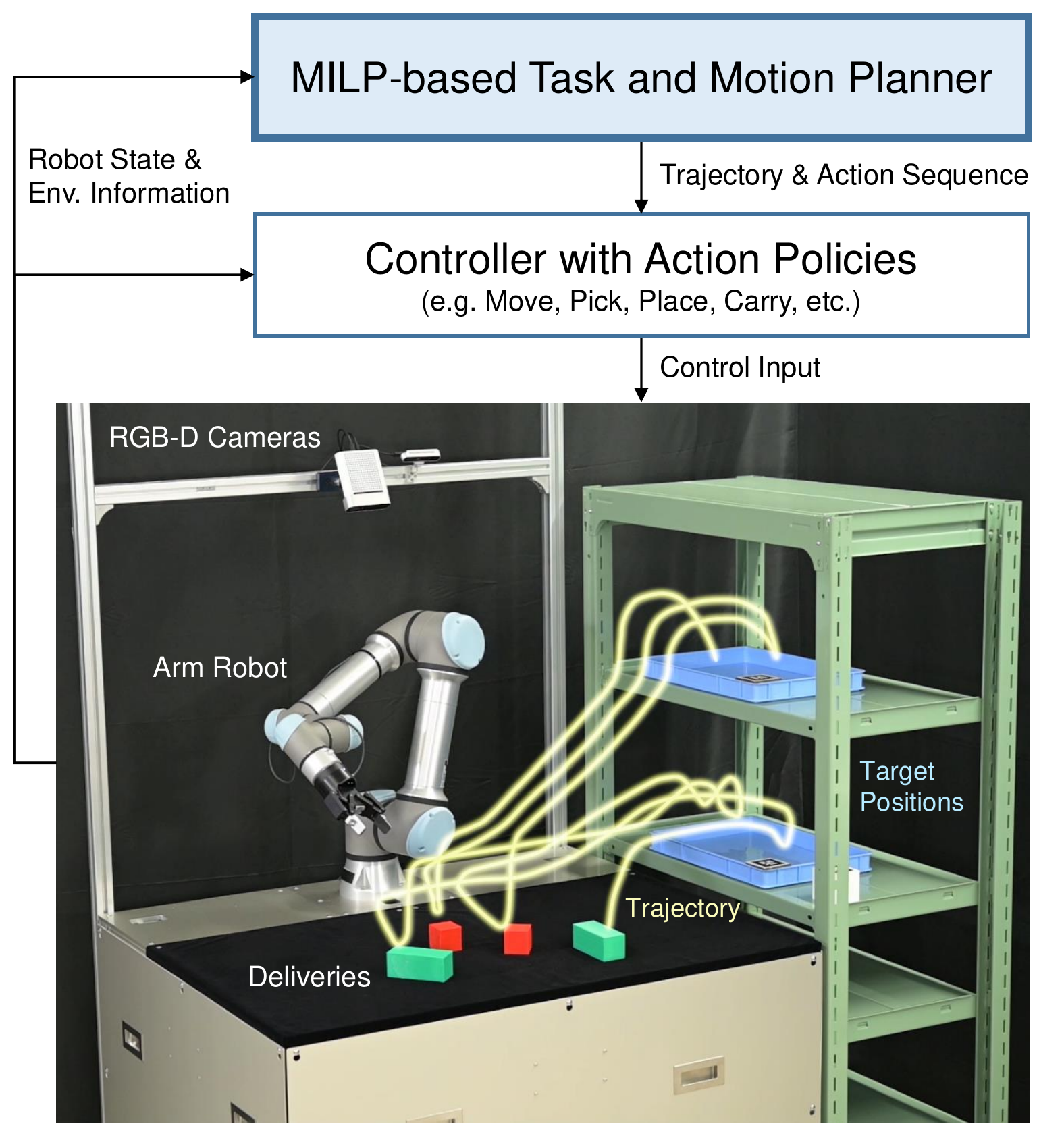}
    \caption{Overview of system architecture. The arm robot delivers target objects (deliveries) to target positions, which are observed by the camera devices. The task and motion planner simultaneously outputs the motion trajectory and action sequence with the settings/observations of the robot and environment. The trajectory consists of discrete time-series data including the positions of the robot’s end-effector (\eg two finger gripper, vacuum hand, etc.). The action sequence consists of discrete time-series of symbolic actions (e.g. Move, Pick, Place, Carry, etc.) grounded in the mode transitions of system dynamics. The action policies are pre-designed for each symbolic action. The controller calculates the control input of the robot by interpolating the trajectory and converting the action sequence to action policies simultaneously.} 
    \label{fig:sys_arch}
    \vspace{-5mm}
\end{figure}

As a {\it teaching-less} approach, task and motion planning (TAMP) is a promising method \cite{c02}--\cite{c09}. TAMP can automatically generate action sequences and motion trajectory of robotic P\&P with given information about the robot and workspace (Fig. \ref{fig:sys_arch}). TAMP provides an advantage to robotic P\&P consisting of discrete action sequences and continuous motion because it directly solves both of the plannings in a cooperative manner. Therefore, TAMP works especially for complex situations where non-trivial dependencies exist between the action sequences and motion trajectory.

Existing approaches for TAMP can be classified into two types: hierarchical feedback \cite{c02}--\cite{c05} and optimization-based \cite{c06}--\cite{c09}. In the hierarchical feedback approach \cite{c02}--\cite{c05}, task planning and motion planning are alternately computed and provide feedbacks to each other to resolve infeasibility until a feasible motion plan is found. Although this approach has shown promising results, it requires parameter tuning on alternation control when applying it to various environments. In the optimization-based approach \cite{c06}--\cite{c09}, TAMP is formulated as optimization models where a solution consists of both task and motion plans. This approach comparatively requires less parameter tuning for various environments because the optimization model is systematically designed and solved. However, the optimization-based TAMP requires a high computation cost for solving when a system's scale and/or complexity increases (\eg collision avoidance).

%%%%%%%%%%%%%%%%%%%%%%%%%%%%%%%%%%%%%%%%%%%%%%%%%%%%%%%%%%%%%%%%%%%%%%%%%%%%%%%%

\subsection{Contributions}
In this paper, we propose new models of optimization-based TAMP for robotic P\&P, which plan action sequences and collision-free motion trajectory with low computation cost. Our motivation is reducing the computation cost of the optimization model based on the state-of-the-art TAMP \cite{c08} integrated with the state-of-the-art collision avoidance \cite{c11}, which is formulated as a mixed-integer linear programming (MILP) problem. To the best of our knowledge, such an integrated model of MILP-based TAMP has not been fully studied in application to robotic P\&P. Therefore, we propose two approaches leveraging features of collision avoidance in robotic P\&P to reformulate the integrated model. Our contribution is presenting two models implemented with the approaches and their effectiveness as follows:
\begin{itemize}
    \item The first proposed model reduces the number of binary variables related to the collision avoidance of deliveries. In this model, the binary variables are successfully reformulated as continuous variables with additional hard constraints. These hard constraints propagate the binary values, which are related to the carry action state and collision avoidance of robots, to the reformulated continuous variables.
    \item The second proposed model enables the MILP solver to find integer solutions more efficiently. In this model, a new term is added to the objective function as a soft constraint which softly restricts the robot's routes around a delivery object. This soft constraint guides the binary variables, which are related to collision avoidance with deliveries, to be fixed to the binary values in the MILP solver's search process. 
\end{itemize}
%%%%%%%%%%%%%%%%%%%%%%%%%%%%%%%%%%%%%%%%%%%%%%%%%%%%%%%%%%%%%%%%%%%%%%%%%%%%%%%%
%%%%%%%%%%%%%%%%%%%%%%%%%%%%%%%%%%%%%%%%%%%%%%%%%%%%%%%%%%%%%%%%%%%%%%%%%%%%%%%%
%%%%%%%%%%%%%%%%%%%%%%%%%%%%%%%%%%%%%%%%%%%%%%%%%%%%%%%%%%%%%%%%%%%%%%%%%%%%%%%%

\section{Related Work}
\label{sec:related}
%%%%%%%%%%%%%%%%%%%%%%%%%%%%%%%%%%%%%%%%%%%%%%%%%%%%%%%%%%%%%%%%%%%%%%%%%%%%%%%%

\subsection{Task and Motion Planning}
 Optimization-based TAMPs have been proposed \cite{c06}--\cite{c09}, and they are modeled as mathematical programming problems (\ie mixed-integer nonlinear programming \cite{c06}, \cite{c07}, MILP \cite{c08}, and continuous nonlinear programming \cite{c09}). The MILP-based TAMP proposed in \cite{c08} is formulated with constraints consisting of task conditions (\eg completion, safety, etc.) and robot dynamics approximated with a low-dimensional mixed logical dynamical system. To achieve a plan in a low-dimensional space, pre-designed action policies are selected at a low-level control. We focus on this MILP-based TAMP as a promising method because optimization solvers have been developed to a practical level (See \ref{subsec:milp}). However, further study is needed for collision avoidance because it is not fully considered in the prior work.

%%%%%%%%%%%%%%%%%%%%%%%%%%%%%%%%%%%%%%%%%%%%%%%%%%%%%%%%%%%%%%%%%%%%%%%%%%%%%%%%

\subsection{Collision Avoidance}
Much research has been conducted on collision avoidance. Many major approaches have been proposed and developed, such as potential field, random sampling, graph search, reachability analysis, optimization, etc. \cite{c10}, \cite{c11}, especially in the field of mobile robots. Collision avoidance problems are generally NP-hard, and practical methods basically rely on heuristics for specific problems \cite{c10}. As for safety integrity, optimization-based collision-free methods have been proposed and applied \cite{c11}--\cite{c13}. These methods completely eliminate the cut-through issue caused by discrete-time models. In this paper, we focus on state-of-the-art collision-free encoding \cite{c11} (Fig. \ref{fig:collision}) as a promising method because it is compatible with MILP-based TAMP \cite{c08}. To the best of our knowledge, MILP-based TAMP integrated with collision-free formulation has not been fully studied, especially regarding computation efficiency. For all of these reasons, we address specific issues and leverage features of collision avoidance in the MILP-based TAMP of robotic P\&P.

%%%%%%%%%%%%%%%%%%%%%%%%%%%%%%%%%%%%%%%%%%%%%%%%%%%%%%%%%%%%%%%%%%%%%%%%%%%%%%%%
\subsection{MILP Solver}
\label{subsec:milp}
\begin{figure}[!t]
    \centering
    \includegraphics[keepaspectratio,clip,width=0.81\hsize]{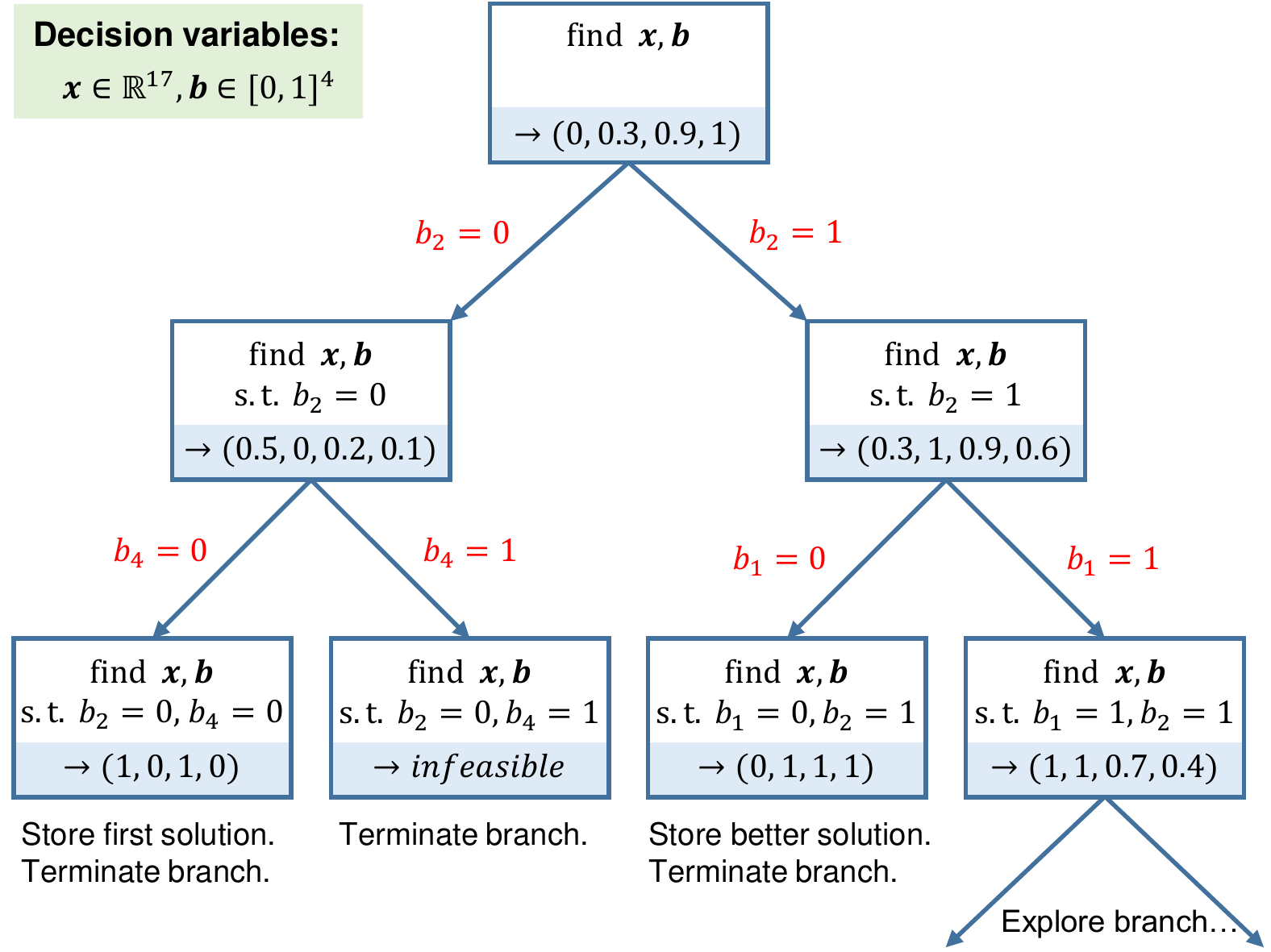}
    \vspace{-3mm}
    \caption{Search example of B\&B method. The basic strategy is a brute-force tree search recursively fixing binary variables one-by-one ({\it branching}) with early termination on the basis of theoretical conditions ({\it bounding}). In the B\&B method, the continuous relaxation problem (CRP), where binary variables of the original MILP problem are converted to continuous 0-1 variables, is solved by linear programming. In a tree-search manner, the CRP is recursively updated with an additional constraint on a single original binary variable (highlighted in red) by using a fractional solution (shaded in blue) of the parent CRP. An incumbent solution is updated with a better integer solution until no more branches exist. {\it Branching} is terminated when any of the following conditions are satisfied: 1) an integer solution is found, 2) a CRP is infeasible, 3) a fractional solution is worse than the incumbent solution, and 4) a MIP gap, which is the difference between the objective values of the best and bound, is smaller than the setting value.}
    \label{fig:B&B}
    \vspace{-4mm}
\end{figure}

MILP is becoming common to solve real-world problems owing to the following contributions: nonlinearity can be reasonably approximated to MILP by using well-known formulation techniques (\eg big-M method), and MILP solvers have been over 3000 times algorithmically faster since 1988 \cite{c14}. Reports on modern commercial MILP solvers have also shown consistent speed-up and expansion of solvable problems since 2010 \cite{c15}, \cite{c16}. However, MILP is NP-hard in general, and it requires a high computation cost when formulation is not appropriate for the branch-and-bound (B\&B) method (Fig. \ref{fig:B&B}). The B\&B method is a fundamental algorithm for most modern MILP solvers \cite{c17}. Therefore, the following two aspects, which are empirically proven to impact search efficiency of B\&B-based MILP solvers, are considered in the practice of formulation: {\it size} (\eg number of binary variables) and {\it tightness} (\eg feasible space of continuous relaxation problem) \cite{c17}, \cite{c18}. We address both {\it size} and {\it tightness} with leveraging features of collision avoidance in the MILP-based TAMP of robotic P\&P.

%%%%%%%%%%%%%%%%%%%%%%%%%%%%%%%%%%%%%%%%%%%%%%%%%%%%%%%%%%%%%%%%%%%%%%%%%%%%%%%%
%%%%%%%%%%%%%%%%%%%%%%%%%%%%%%%%%%%%%%%%%%%%%%%%%%%%%%%%%%%%%%%%%%%%%%%%%%%%%%%%
%%%%%%%%%%%%%%%%%%%%%%%%%%%%%%%%%%%%%%%%%%%%%%%%%%%%%%%%%%%%%%%%%%%%%%%%%%%%%%%%

\begin{figure*}[!t]
    \centering
    \includegraphics[keepaspectratio,clip,width=\hsize]{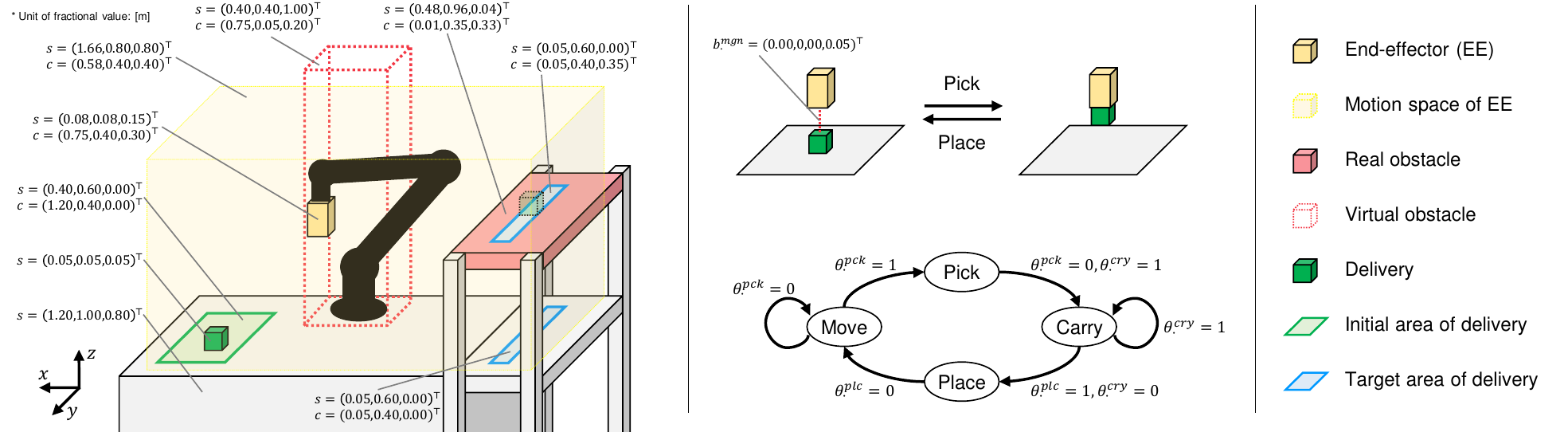}
    \vspace{-6mm}
    \caption{Model of P\&P planning. The right part shows the symbols of all modeled instances in this figure. The left part shows an example configuration of a P\&P system with the instances and their parameters (shape width $s$ and center position $c$). This configuration is also used in the evaluation. The following three simplifications are assumed to fit the computation cost within a practical range: 1) the instances are modeled as axis-aligned bounding boxes (AABB), 2) the dynamics of a robot's body part (black object) is not modeled, 3) irrelevant obstacles (gray objects) outside the motion space of the end-effector are not modeled. The center part shows state transitions of the delivery (top) and the end-effector (bottom) regarding the actions. The end-effector and grasping delivery face each other on the bottom and top surfaces, respectively. Moreover, the end-effector takes margin $b^{mgn}_{\bullet}$ between itself and the grasping delivery at the pick/place actions.}
    \label{fig:model}
    \vspace{-4mm}
\end{figure*}

\section{Baseline Model}
%%%%%%%%%%%%%%%%%%%%%%%%%%%%%%%%%%%%%%%%%%%%%%%%%%%%%%%%%%%%%%%%%%%%%%%%%%%%%%%%

\subsection{System Architecture}
Fig. \ref{fig:sys_arch} shows an overview of our system architecture. Our system architecture is based on prior studies \cite{c08}, \cite{c09} and it has the only difference in regard to the model of the task and motion planner. In this paper, we assumed a single arm robot configuration for a simple explanation and evaluation. Note, however, that multi-arm robot configurations are also supported by the model of the task and motion planner as formulated in the following subsections \ref{subsec:notation} and \ref{subsec:formulation}.

%%%%%%%%%%%%%%%%%%%%%%%%%%%%%%%%%%%%%%%%%%%%%%%%%%%%%%%%%%%%%%%%%%%%%%%%%%%%%%%%

\subsection{Modeling Settings}
In this paper, we introduce our baseline model of the task and motion planner for a robotic P\&P. Our baseline model is based on the MILP-based TAMP of \cite{c08}. Moreover, our baseline model is newly integrated with collision-free encoding \cite{c11} for safer collision avoidance. In addition, our baseline model has the following minor improvements: 
\begin{itemize}
    \item Simple formulation of delivery's dynamics with an approach point (Eqs. (7)--(13), center of Fig. \ref{fig:model}).
    \item Strengthened formulation for search efficiency of MILP solver (Eqs. (16)--(17)).
    \item Direct minimization of time steps to complete P\&P with consideration of  distance minimization (Eqs. (26)--(30)).
\end{itemize}

Fig. \ref{fig:model} illustrates our baseline model formulated as a MILP problem. The purpose of the baseline model is to generate motion trajectory and action sequences of the end-effector by solving it. The concrete formulation with the constraints and objective function is explained in the following subsections.

\subsection{Notation}
\label{subsec:notation}
$N_{stp}$, $N_{ee}$, $N_{dlv}$, $N_{obs}$, and $N_{rg}$, where $N_{\bullet}\in\mathbb{N}_{+}$ are constants, denote the number of time steps, end-effectors, deliveries, obstacles, and collision-free regions, respectively. Let $\bullet$ denote any string. $t$, $i$, $j_{\bullet}$, $k$, and $r$ respectively represent the following indices: ``$t$-th time step'', ``$i$-th end-effector'', ``$j_{\bullet}$-th delivery'', ``$k$-th obstacle'', and ``$r$-th collision-free region''. Moreover, we define their indices as $t\in\{0,...,N_{stp}\}$, $i\in\{1,...,N_{ee}\}$, $j_{\bullet}\in\{1,...,N_{dlv}\}$, $k\in\{1,...,N_{obs}\}$, and $r\in\{1,..,N_{rg}\}$ where $j_{1} \neq j_{2}$. We give all possible combinations of the indices for each constraint and objective function. The upper/lower bounds of $\bullet$ are constants denoted by $\overline{\bullet}$ and $\underline{\bullet}$, respectively. The initial/target values of $\bullet_{t}$ are also constants denoted by $\bullet_{0}$ and $\bullet_{\star}$, respectively. Furthermore, the relative value of $\bullet_{t}$ from its initial value is denoted by $\bullet^{'}_{t}$ (\ie $p^{'}_{t} = p_{t} - p_{0}$). For simple formulation, we use the logical operators ``NOT'' $\lnot$, ``AND'' $\land$, and ``OR'' $\lor$, which are transformed to the equivalent MILP formulation with additional decision variables as shown in Eqs. (1)--(3): 
\begin{align}   
    \hspace{-0mm} &\phi=\lnot \varphi              \hspace{-3mm} &\Leftrightarrow & \SP \phi = 1 - \varphi, & \\
    \hspace{-0mm} &\phi=\bigwedge_{n}{\varphi_{n}} \hspace{-3mm} &\Leftrightarrow &
        \begin{cases}
            \phi \leq \varphi_{n} \\
            \phi \geq \sum_{n}{\varphi_{n}} -N+1
        \end{cases}
        \hspace{-5mm} & n\in\{1,...,N\},\\
    \hspace{-0mm} &\phi=\bigvee_{n}{\varphi_{n}}   \hspace{-3mm} &\Leftrightarrow &
        \begin{cases}
            \phi \geq \varphi_{n} \\
            \phi \leq \sum_{n}{\varphi_{n}}
        \end{cases}
        \hspace{-5mm} & n\in\{1,...,N\},
\end{align}
where $\phi \in [0,1]$ is the additional decision variable and $\varphi_{n} \in \{0,1\}$ is the operand ($\varphi_{n} \in [0,1]$ is allowable).

\subsection{Formulation}
\label{subsec:formulation} 
Equations (4)--(29) are the constraints of the baseline model. Equation (30) is the objective function of the baseline model. The baseline model is a discrete bounded time system where the time step size $\Delta t$ and its number $N_{stp}$ are respectively given as constant values.

Constraint (4) represents the dynamics of the end-effector:
\begin{align}
    p^{ee}_{i,t+1} = p^{ee}_{i,t} + v^{ee}_{i,t} \Delta t
\end{align}
where $p^{ee}_{i,t},v^{ee}_{i,t} \in \mathbb{R}^{3}$ are the position and velocity of the end-effector, respectively. For simplification, we use such a discrete state-space model where the control input is velocity as the dynamics of the end-effector.

Constraints (5) and (6) represent the feasible interaction among the end-effectors and deliveries, which means ``one end-effector can grasp one delivery at the same time'':
\begin{align}   
    \sum_{i}{\theta^{gsp}_{i,j,t}} \leq 1&, \\
    \sum_{j}{\theta^{gsp}_{i,j,t}} \leq 1&
\end{align}

where $\theta^{gsp}_{i,j,t} \in \{0,1\}$ is the action state where the end-effector is grasping the delivery.

Constraints (7)--(9) represent the states corresponding to the action sequences of the end-effector:
\begin{align}
    &\theta^{pck}_{i,j,t} = (\lnot\theta^{gsp}_{i,j,t}) \wedge \theta^{gsp}_{i,j,t+1}, \\
    &\theta^{plc}_{i,j,t} = (\lnot\theta^{gsp}_{i,j,t}) \wedge \theta^{gsp}_{i,j,t-1}, \\
    &\theta^{cry}_{i,j,t} = \theta^{gsp}_{i,j,t} - \theta^{pck}_{i,j,t}
\end{align}
where $\theta^{pck}_{i,j,t}, \theta^{plc}_{i,j,t}, \theta^{cry}_{i,j,t} \in [0,1]$ are the action states where the end-effector is picking, placing, and carrying the delivery, respectively (See Fig. \ref{fig:model}). These constraints decompose and propagate the binary values of $\theta^{gsp}_{i,j,t}$ to $\theta^{pck}_{i,j,t}$, $\theta^{plc}_{i,j,t}$, and $\theta^{cry}_{i,j,t}$.

Constraints (10)--(13) represent the dynamics of the delivery as shown in Fig. \ref{fig:model}:
\begin{align}
    -M_{1}(\lnot\theta^{cry}_{i,j,t}) \leq p^{dlv}_{j,t} - p^{ee}_{i,t} + b^{on}_{i,j}\;\: \leq M_{1}(\lnot\theta^{cry}_{i,j,t}),\\
    -M_{2}(\lnot\theta^{pck}_{i,j,t}) \leq p^{dlv}_{j,t} - p^{ee}_{i,t} + b^{off}_{i,j} \leq M_{2}(\lnot\theta^{pck}_{i,j,t}),\\
    -M_{3}(\lnot\theta^{plc}_{i,j,t}) \leq p^{dlv}_{j,t} - p^{ee}_{i,t} + b^{off}_{i,j} \leq M_{3}(\lnot\theta^{plc}_{i,j,t}), \\
    -M_{4}\sum_{i}\theta^{gsp}_{i,j,t} \leq p^{dlv}_{j,t+1} - p^{dlv}_{j,t}\SPP\SP \leq M_{4}\sum_{i}\theta^{gsp}_{i,j,t}
\end{align}
where $p^{dlv}_{j,t} \in \mathbb{R}^{3}$ is the position of the delivery. Moreover, $b^{on}_{i,j},b^{off}_{i,j} \in \mathbb{R}^{3}$ are constants representing the relative position between the delivery and the end-effector when the delivery is carried and not carried, respectively. The concrete values of the constants are given by $b^{on}_{i,j} = (0,0,\frac{1}{2})^{\top}\circ(s^{ee}_{i}+s^{dlv}_{j})$, $b^{off}_{i,j}=b^{on}_{i,j}+b^{mgn}_{i,j}$ where $\circ$ is the operator of an element-wise product and  $s^{ee}_{i},s^{dlv}_{j} \in \mathbb{R}^{3}_{+}$ are constants representing the shape width of the end-effector and delivery, respectively. $M_{1},...,{M_{4}}$ are large coefficients to switch dynamics via inequality constraints with binary variables, and we give the tightest value to each coefficient of the constraints.

Constraints (14) and (15) represent the P\&P completion conditions for each of the deliveries:
%\begin{align}    
\begin{gather}
    -M_{5}(\lnot\psi_{j,t}) \leq p^{dlv}_{j,t} - p^{dlv}_{j,\star} \leq M_{5}(\lnot\psi_{j,t}),\\
    \psi_{j,t} \leq \lnot\sum_{i}{\theta^{gsp}_{i,j,t}}
\end{gather}
%\end{align}
where $\psi_{j,t} \in \{0,1\}$ is the state where the delivery is at its target position without being grasped, and $M_{5}$ is a large coefficient similar to $M_{1},...,M_{4}$.

Constraints (16) and (17) force $\psi_{j,t}$ to take 1 when the delivery is at its target position by preventing $\psi_{j,t}$ from taking 0. As a result, the computation cost of the MILP solver can be substantially reduced. Constraint (16) means that the delivery must be at its target position when grasping is finished. Constraint (17) means that the delivery keeps its position once it reached its target position.

%\begin{align}
\begin{gather}
    \psi_{j,t+1} - \psi_{j,t} \geq \sum_{i}{(\theta^{gsp}_{i,j,t}-\theta^{gsp}_{i,j,t+1})}, \\[-3pt]
    \psi_{j,0} \leq \psi_{j,1} \leq...\leq \psi_{j,N_{stp}} = 1.
\end{gather}
%\end{align}

\begin{figure}[!t]
    \centering
    \includegraphics[keepaspectratio,clip,width=\hsize]{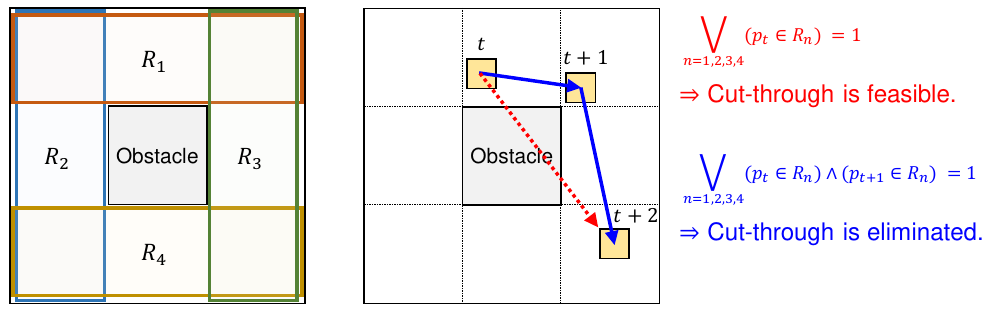}
    \vspace{-6mm}    
    \caption{Model of collision avoidance \cite{c11}. The left part shows a 2D space example of a collision-free region, which is defined as AABB on each side of the obstacle (\ie $R_{1},...,R_{4}$). The important point is that each collision-free region overlaps with adjacent regions (\eg $R_{2}$ overlaps with $R_{1}$ and $R_{4}$). The right part shows a collision-free path and constraint (highlighted in blue) compared with the case of cutting through the obstacle (highlighted in red). The bottom constraint requires that the moving object is positioned within any one of the same collision-free regions at consecutive time steps. This constraint completely eliminates the cut-through of corners.}
    \label{fig:collision}
    \vspace{-5mm}
\end{figure}

Constraints (18)--(29) represent the collision-free encoding \cite{c11} as shown in Fig. \ref{fig:collision}. The collision avoidance of the end-effector with the obstacle is modeled by constraints (18) and (22), and that with the delivery is modeled by constraints (19) and (23). The collision avoidance of the delivery with the obstacle is modeled by constraints (20) and (24), and that with the other delivery is modeled by constraints (21) and (25). On the basis of the big-M method, constraints (18)--(21) evaluate whether a moving object (\ie end-effector and delivery) is within the collision-free region:

\begin{align}
    p^{ee}_{i,t} \leq \overline {R^{obs}_{k,r}} z^{ee,obs}_{i,k,r,t} + \overline {p^{ee}_{i,t}}(\lnot z^{ee,obs}_{i,k,r,t})&,\notag\\
    p^{ee}_{i,t} \geq \underline{R^{obs}_{k,r}} z^{ee,obs}_{i,k,r,t} + \underline{p^{ee}_{i,t}}(\lnot z^{ee,obs}_{i,k,r,t})&
\end{align}
where $R^{obs}_{k,r} \in \mathbb{R}^{3}$ is the position of the collision-free region on the side of the obstacle and $z^{ee,obs}_{i,k,r,t} \in \{0,1\}$ is the state where the end-effector is within the corresponding collision-free region on the side of the obstacle.
\begin{align}
    p^{ee}_{i,t} - p^{dlv'}_{j,t} \leq \overline {R^{dlv}_{j,r}} z^{ee,dlv}_{i,j,r,t} + (\overline {p^{ee}_{i,t} - p^{dlv'}_{j,t}})(\lnot z^{ee,dlv}_{i,j,r,t})&,\notag\\
    p^{ee}_{i,t} - p^{dlv'}_{j,t} \geq \underline{R^{dlv}_{j,r}} z^{ee,dlv}_{i,j,r,t} + (\underline{p^{ee}_{i,t} - p^{dlv'}_{j,t}})(\lnot z^{ee,dlv}_{i,j,r,t})&
\end{align}
where $R^{dlv}_{j,r} \in \mathbb{R}^{3}$ is the position of the collision-free region on the side of the delivery and $z^{ee,dlv}_{i,j,r,t} \in \{0,1\}$ is the state where the end-effector is within the corresponding collision-free region on the side of the delivery.
\begin{align}
    p^{dlv}_{j,t} \leq \overline {R^{obs}_{k,r}} z^{dlv,obs}_{j,k,r,t} + \overline {p^{dlv}_{j,t}}(\lnot z^{dlv,obs}_{j,k,r,t})&,\notag\\
    p^{dlv}_{j,t} \geq \underline{R^{obs}_{k,r}} z^{dlv,obs}_{j,k,r,t} + \underline{p^{dlv}_{j,t}}(\lnot z^{dlv,obs}_{j,k,r,t})&
\end{align}
where $z^{dlv,obs}_{j,k,r,t} \in \{0,1\}$ is the state where the delivery is within the collision-free region on the side of the obstacle.

\begin{align}
    p^{dlv}_{j_{1},t} - p^{dlv'}_{j_{2},t} \leq \overline {R^{dlv}_{j_{2},r}} z^{dlv,dlv}_{j_{1},j_{2},r,t} + (\overline {p^{dlv}_{j_{1},t} - p^{dlv'}_{j_{2},t}})(\lnot z^{dlv,dlv}_{j_{1},j_{2},r,t})&,\notag\\
    p^{dlv}_{j_{1},t} - p^{dlv'}_{j_{2},t} \geq \underline{R^{dlv}_{j_{2},r}} z^{dlv,dlv}_{j_{1},j_{2},r,t} + (\underline{p^{dlv}_{j_{1},t} - p^{dlv'}_{j_{2},t}})(\lnot z^{dlv,dlv}_{j_{1},j_{2},r,t})&.
\end{align}
where $z^{dlv,dlv}_{j_{1},j_{2},r,t} \in \{0,1\}$ is the state where the delivery is within the collision-free region on the side of the other obstacle.

Constraints (22)--(25) force the moving objects to be within any one of the same collision-free regions at two consecutive time steps to completely prevent cut-through:
\begin{align}
    & \bigvee_{r}{(z^{ee,obs} _{i,k,r,t}        \land z^{ee,obs}_{i,k,r,t+1})}          \hspace{-20mm}&= 1, \\
    & \bigvee_{r}{(z^{ee,dlv} _{i,j,r,t}        \land z^{ee,dlv}_{i,j,r,t+1})}          \hspace{-20mm}&= 1, \\
    & \bigvee_{r}{(z^{dlv,obs}_{j,k,r,t}        \land z^{dlv,obs}_{j,k,r,t+1})}         \hspace{-20mm}&= 1, \\
    & \bigvee_{r}{(z^{dlv,dlv}_{j_{1},j_{2},r,t}\land z^{dlv,dlv}_{j_{1},j_{2},r,t+1})} \hspace{-20mm}&= 1.
\end{align}

Constraints (26) and (27) are for exploiting the completion time steps of all P\&P and moving distances of the end-effectors to be used in the objective function:
%\begin{align}
\begin{gather}
    C_{t} = \bigwedge_{\tau=t,...,N_{stp}}\bigwedge_{j}{\psi_{j,\tau}},\\
    - u_{i,t} \leq v^{ee}_{i,t} \leq u_{i,t}
\end{gather}
%\end{align}
where $C_{t}\in [0,1]$ is the state where the P\&P of all deliveries is completed and $u_{i,t} \in \mathbb{R}^{3}_{\geq 0}$ is a slack variable to obtain the absolute value of $v^{ee}_{i,t}$ by combining it with the objective function (\ie minimizing $u_{i,t}$).

Constraints (28) and (29) represent the terms of the objective function.
\begin{align}
    &J_{time} =\frac{1}{N_{stp}+1} \sum_{t}(\lnot C_{t}), \\
    &J_{dist} =\sum_{t}\sum_{i}{w_{t}\|u_{i,t}\|_{1}}
\end{align}
where $J_{time}$ is the normalized time steps to complete all P\&P and $J_{dist}$ is the weighted sum of the L1 distance where the end-effectors move. $w_{t} \in \mathbb{R}_{+}$ is a small weight coefficient to prioritize $J_{time}$ over $J_{dist}$ to evaluate minimality. We give conservative values to $w_{t}$ by $w_{t}=(1+\alpha)^{\frac{t}{N_{stp}}-1} / ((N_{stp}+1)^2\sum_{i}\overline{\|v^{ee}_{i,t}\|_{1}})$ where $\alpha \in \mathbb{R}_{+}$ is a penalty coefficient. These $w_{t}$ guarantee $\frac{1}{N_{stp}+1} \geq J_{dist}$ where the minimum variation of $J_{time}$ is greater than the maximum variation of $J_{dist}$. In addition, these $w_{t}$ contribute to generate reasonable motion trajectory because moves are penalized at later time steps. 

Equation (30) represents the objective function of the baseline model:
\begin{align}
    X^{*} = & \argmin_{X}{(J_{time}+J_{dist})} \\ \nonumber
            & \st Eq. \; (4)\text{--}(29)
\end{align}
where $X^{*}$ is the optimal solution of $X$ which is a vector consisting of all decision variables (\eg $p^{ee}_{i,t}$, $v^{ee}_{i,t}$, $\theta^{pck}_{i,j,t}$, $\theta^{plc}_{i,j,t}$, $\theta^{cry}_{i,j,t}$, etc.).

The model can be solved by the MILP solver. Therefore, the optimal $p^{ee}_{i,t}$, $v^{ee}_{i,t}$, $\theta^{pck}_{i,j,t}$, $\theta^{plc}_{i,j,t}$ and $\theta^{cry}_{i,j,t} $ can be extracted from $X^{*}$ as the trajectory and actions sequences.

%%%%%%%%%%%%%%%%%%%%%%%%%%%%%%%%%%%%%%%%%%%%%%%%%%%%%%%%%%%%%%%%%%%%%%%%%%%%%%%%
%%%%%%%%%%%%%%%%%%%%%%%%%%%%%%%%%%%%%%%%%%%%%%%%%%%%%%%%%%%%%%%%%%%%%%%%%%%%%%%%
%%%%%%%%%%%%%%%%%%%%%%%%%%%%%%%%%%%%%%%%%%%%%%%%%%%%%%%%%%%%%%%%%%%%%%%%%%%%%%%%
\section{Proposed Approach}

%%%%%%%%%%%%%%%%%%%%%%%%%%%%%%%%%%%%%%%%%%%%%%%%%%%%%%%%%%%%%%%%%%%%%%%%%%%%%%%%

\subsection{Motivation and Overview}
Our motivation is to reduce the computation cost of the baseline model via leveraging features of collision avoidance in robotic P\&P. To the best of our knowledge, few prior studies have addressed such work. We take two approaches improving the {\it size} and {\it tightness} of the baseline model, which impact the search efficiency of the MILP solver as mentioned in subsection \ref{subsec:milp}. The purpose of the first approach regarding {\it size} is to reduce the search space of the MILP solver. The purpose of the second approach regarding {\it tightness} is to make {\it bounding} work more efficiently, in other words, to guide the MILP solver to find integer solutions at shallower nodes in the search tree. The policy of the first approach is to reduce the number of binary variables in regard to collision avoidance of the deliveries. The policy of the second approach is to softly tighten the values of binary variables in regard to collision avoidance with the deliveries. In this paper, we implement these approaches by using continuous relaxation with hard constraints and a soft constraint, respectively. Table \ref{tab:comparison} shows a detailed comparison of the baseline and proposed models. ``Ours (hard)'' is based on the first approach, and ``Ours (hard+soft)'' is based on both of the first and second approaches.
 
\begin{table}[!t]
    \centering
    \caption{comparison of models}
    \vspace{-2mm}
    \begin{tabular}{|l|c|c|c|} \hline
        \multicolumn{1}{|c|}{\bf Model} & {\bf Variable} & {\bf Constraints} & \!{\bf Objective value}\! \\ 
                         & {\tiny $ z^{dlv,\bullet}_{\bullet} \in ...$} & \!{\fontsize{5.75pt}{0pt}\selectfont Eq. (4)--(23),(26)--(29),...}\! & {\tiny $J_{time}+J_{dist}...$}\\ \hline
        Baseline           & $\{0,1\}$ & Eq. (24), (25) &  \\ 
        Ours (hard)        & $[0,1]$   & Eq. (31), (32) &  \\ 
        Ours (hard+soft)\! & $[0,1]$   & Eq. (31)--(33) & $+J_{route}$ \\ \hline      
    \end{tabular}
    \label{tab:comparison}
    \vspace{-4mm}
\end{table}
%%%%%%%%%%%%%%%%%%%%%%%%%%%%%%%%%%%%%%%%%%%%%%%%%%%%%%%%%%%%%%%%%%%%%%%%%%%%%%%%

\subsection{Hard Constraint}
For the first approach, we introduced an assumption where the collision-free regions of the end-effector and carried delivery are identified. This assumption can be simply formulated as the following additional constraints:
\begin{align}
    z^{dlv,obs}_{j,k,r,t}        &= \bigvee_{i}{(z^{ee,obs}_{i,k,r,t} \land \theta^{cry}_{i,j,t})}, \\
    z^{dlv,dlv}_{j_{1},j_{2},r,t}&= \bigvee_{i}{(z^{ee,dlv}_{i,j_{2},r,t} \land \theta^{cry}_{i,j_{1},t})}.
\end{align}
As a result, the values of $z^{dlv,obs}_{j,k,r,t}$ and $z^{dlv,dlv}_{j_{1},j_{2},r,t}$ are conditionally bound with values of $z^{ee,obs}_{i,k,r,t}$ and $z^{ee,dlv}_{i,j_{2},r,t}$, respectively. In other words, $z^{dlv,obs}_{j,k,r,t}$ and $z^{dlv,dlv}_{j_{1},j_{2},r,t}$ can be reformulated as 0-1 continuous variables. For all $r$, their values take 0 when the delivery is carried by no end-effector. This possibly induces an infeasibility of the collision-free encoding, but it can be resolved by removing constraints (24) and (25) from the set of constraints. In this case, the lack of constraints (24) and (25) does not affect the collision avoidance of the delivery because the delivery always stays within any collision-free regions when it is not carried by the end-effectors. This insight also plays an important role of the first approach. Table \ref{tab:comparison} shows a summary of the aforementioned description as ``Ours (hard)''.

%%%%%%%%%%%%%%%%%%%%%%%%%%%%%%%%%%%%%%%%%%%%%%%%%%%%%%%%%%%%%%%%%%%%%%%%%%%%%%%%

\begin{figure*}[!t]
    \centering
    \includegraphics[keepaspectratio,clip,width=\hsize]{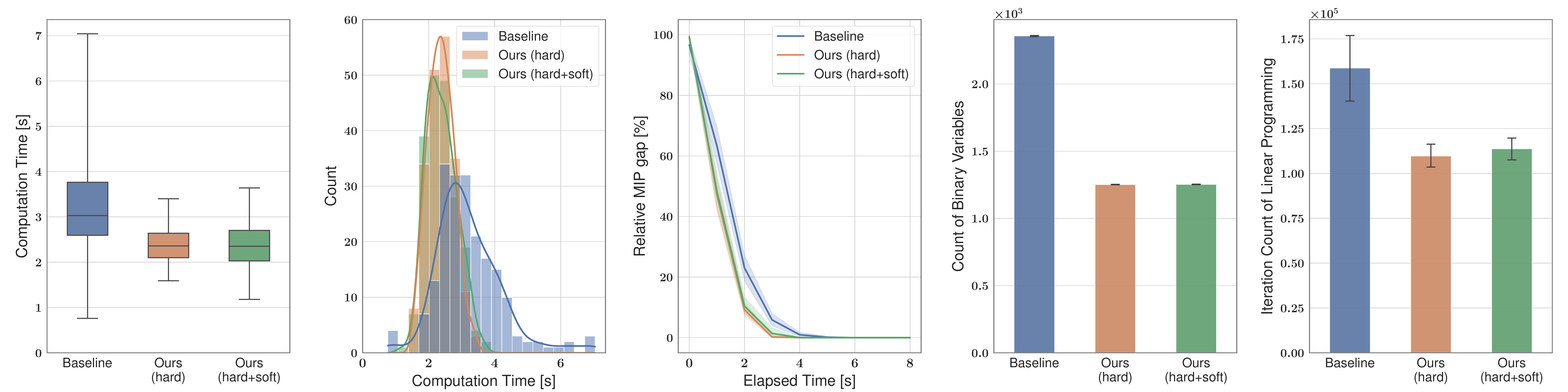}
    \vspace{-6mm}
    \caption{Result of optimization in the case of $N_{dlv}=2$ ($N_{stp}=30$). The graphic representation is as follows: 1) boxplot, 2) histogram (translucent bar) with kernel density estimation (solid line), 3) time series of the mean (solid line) with 99\% confidence intervals (shaded area), and 4), 5) bar chart of the mean (solid bar) with 99\% confidence intervals (black bold bar), respectively. Computation time means the time to find an optimal solution. Binary variables were counted after the models were presolved. Iteration count of linear programming means count of CRPs solved by the MILP solver.}
    \label{fig:result1}
\end{figure*}
\begin{figure*}[!t]
    \centering
    \includegraphics[keepaspectratio,clip,width=\hsize]{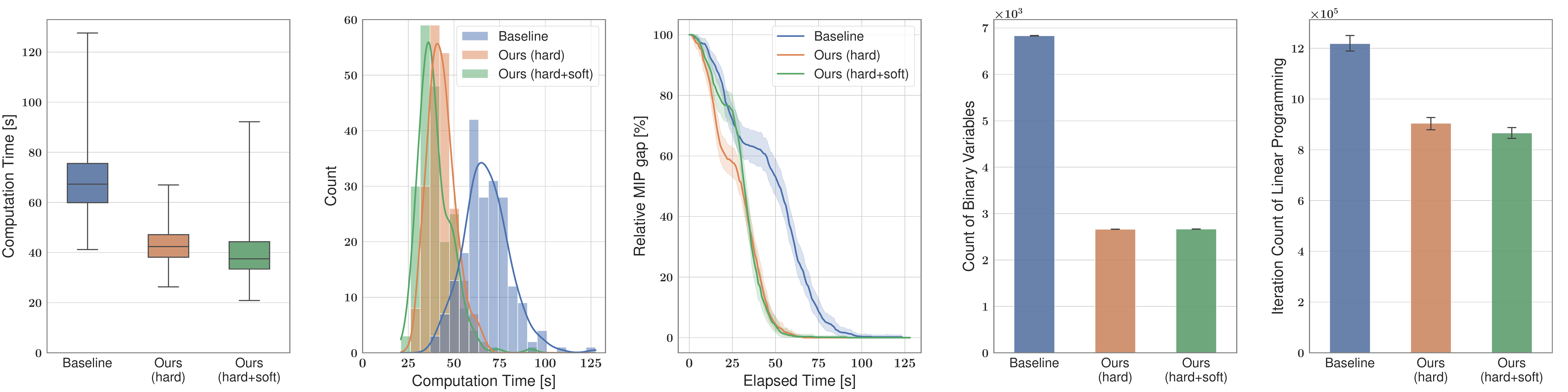}
    \vspace{-6mm}
    \caption{Result of optimization in the case of $N_{dlv}=3$ ($N_{stp}=45$). The graphic representation is in a similar manner with Fig. \ref{fig:result1}.} 
    \label{fig:result2}
    \vspace{-1mm}
\end{figure*}

\subsection{Soft Constraint}
For the second approach, we introduced another assumption where it is feasible to softly restrict the end-effector's routes around the deliveries. The reason we choose such a soft constraint is that it is difficult to preliminarily determine the collision-free regions not to be passed by the end-effector as a hard constraint. The second assumption can be also simply formulated as follows:

\begin{align}
    J_{route} = \sum_{t}\sum_{i}\sum_{j}\sum_{r\in R^{c}_{i,j,t}}{z^{ee,dlv}_{i,j,r,t}},
\end{align}
\begin{align}
    X^{*} = & \argmin_{X}{(J_{time}+J_{dist}+J_{route})} \\ \nonumber 
            & \st Eq. \; (4)\text{--}(23),(26)\text{--}(29),(31)\text{--}(33) 
\end{align}
where $J_{route}$ is the sum of the penalties for the end-effector to pass through the collision-free regions on the sides of delivery, which are selected from $R^{c}_{i,j,t}$. Moreover, $R^{c}_{i,j,t}$ is the set of the collision-free regions around the delivery, which is pre-configured on the basis of the predicted probability of the pass-through. In this paper, we simply configured all $R^{c}_{i,j,t}$ to contain three collision-free regions as follows: the first is on the bottom side of the delivery and the remaining are on both sides of the delivery along the horizontal axis where the end-effector moves less (\ie y-axis in Fig. \ref{fig:model}). This simple way is based on the prior knowledge about routes in robotic P\&P where the end-effector typically moves over the deliveries and along a long axis direction. 
Note that a new term $J_{route}$ is added to the objective function. $J_{route}$ is minimized most preferentially because the minimum variation of $J_{route}$ is greater than the maximum variation of $J_{time}+J_{dist}$. As a result, this soft constraint makes the MILP solver find fractional solutions consisting of more $z^{ee,dlv}_{i,j,r,t}=0$. In other words, the MILP solver has more chances of {\it bounding} where the integer solutions are found. The constraint is inherited from ``Our (hard)'' for the evaluation. Table \ref{tab:comparison} shows a summary of the aforementioned description as ``Ours (hard+soft)''.

%%%%%%%%%%%%%%%%%%%%%%%%%%%%%%%%%%%%%%%%%%%%%%%%%%%%%%%%%%%%%%%%%%%%%%%%%%%%%%%%
%%%%%%%%%%%%%%%%%%%%%%%%%%%%%%%%%%%%%%%%%%%%%%%%%%%%%%%%%%%%%%%%%%%%%%%%%%%%%%%%
%%%%%%%%%%%%%%%%%%%%%%%%%%%%%%%%%%%%%%%%%%%%%%%%%%%%%%%%%%%%%%%%%%%%%%%%%%%%%%%%
\section{Evaluation}

%%%%%%%%%%%%%%%%%%%%%%%%%%%%%%%%%%%%%%%%%%%%%%%%%%%%%%%%%%%%%%%%%%%%%%%%%%%%%%%%

\subsection{Settings}
We compare the baseline model and proposed models as shown in Table \ref{tab:comparison}. All of the models use the same configuration with the parameters shown in Fig. \ref{fig:model} and the remaining parameters are as follows: $\Delta t=0.50$ s, $N_{stp}=15N_{dlv}$, $N_{dlv} \in \{2,3\}$, $\overline{|v^{ee}_{i,t}|} = (0.40,0.20,0.20)^{\top}$ m/s, and  $\alpha=1$. We randomly sampled 200 combinations of initial and target positions of the deliveries, which are within the area shown in Fig. \ref{fig:model}. We solved all of the models implemented by Python-mip 1.13.0 \cite{c19} with the MILP solver Gurobi 9.1.1 \cite{c20} on an Intel Core i7-10700K (8-core/3.80GHz) and 16-GB RAM. We basically used the default settings of Gurobi in 8-thread parallel execution and 300 s time limit, but disabled cut generation and feasibility pumping, which equally slowed computation down for all of the models.

%%%%%%%%%%%%%%%%%%%%%%%%%%%%%%%%%%%%%%%%%%%%%%%%%%%%%%%%%%%%%%%%%%%%%%%%%%%%%%%%

\subsection{Results}
Fig. \ref{fig:result1} shows the result in the case of $N_{dlv}=2$ ($N_{stp}=30$). The boxplot shows that both of the proposed models reduced the computation time by about 23\% in the range of the 25--75th percentiles. Similarly, the histogram shows that both the mean and variance of the computation time are improved. However, there is only a slight difference of the variance compared with that of the mean between the proposed models. The time-series data of the relative MIP gap shows that the proposed models have better convergence properties. The left bar chart shows that the number of binary variables were reduced by about 50\% in the proposed models. The right bar chart shows that the iteration count of linear programming was reduced by about 31\% in the proposed models. As a result, the computation time was reduced by about 23\%.
 We can summarize that our first approach, which reduces the number of binary variables, certainly improved computation time but our second approach did not contribute in the case of $N_{dlv}=2$.

Fig. \ref{fig:result2} shows the result in the case of $N_{dlv}=3$ ($N_{stp}=45$). These graphs show similar findings to those of $N_{dlv}=2$ (Fig. \ref{fig:result1}) and much greater improvement over the baseline model. We examined the cause of the improvement, where the reduction rate of the computation time (about 38, 46\%) was greater than the reduction rate of the iteration count (about 25, 30\%). We found that the number of constraints were also considerably reduced in the proposed models because of the removal of constraints (24) and (25). It is supposed that small {\it size} model resulted in small computation time per iteration and it contributed to much further speed-up. In addition to that, the computation time was improved even more by the second approach. The right bar chart shows greater improvement in search efficiency by ``Ours (hard+soft)'' and this was not observed in the case of $N_{dlv}=2$. It is supposed that this cumulative speed-up can be larger when $N_{dlv}$ is larger. According to this result and the boxplot, the proposed models independently address the computation efficiency of collision avoidance with deliveries. We can summarize that our second approach, which softly tightens binary variables, further reduces computation time when $N_{dlv}$ is larger.

\begin{figure}[!t]
    \centering
    \includegraphics[keepaspectratio,clip,width=\hsize]{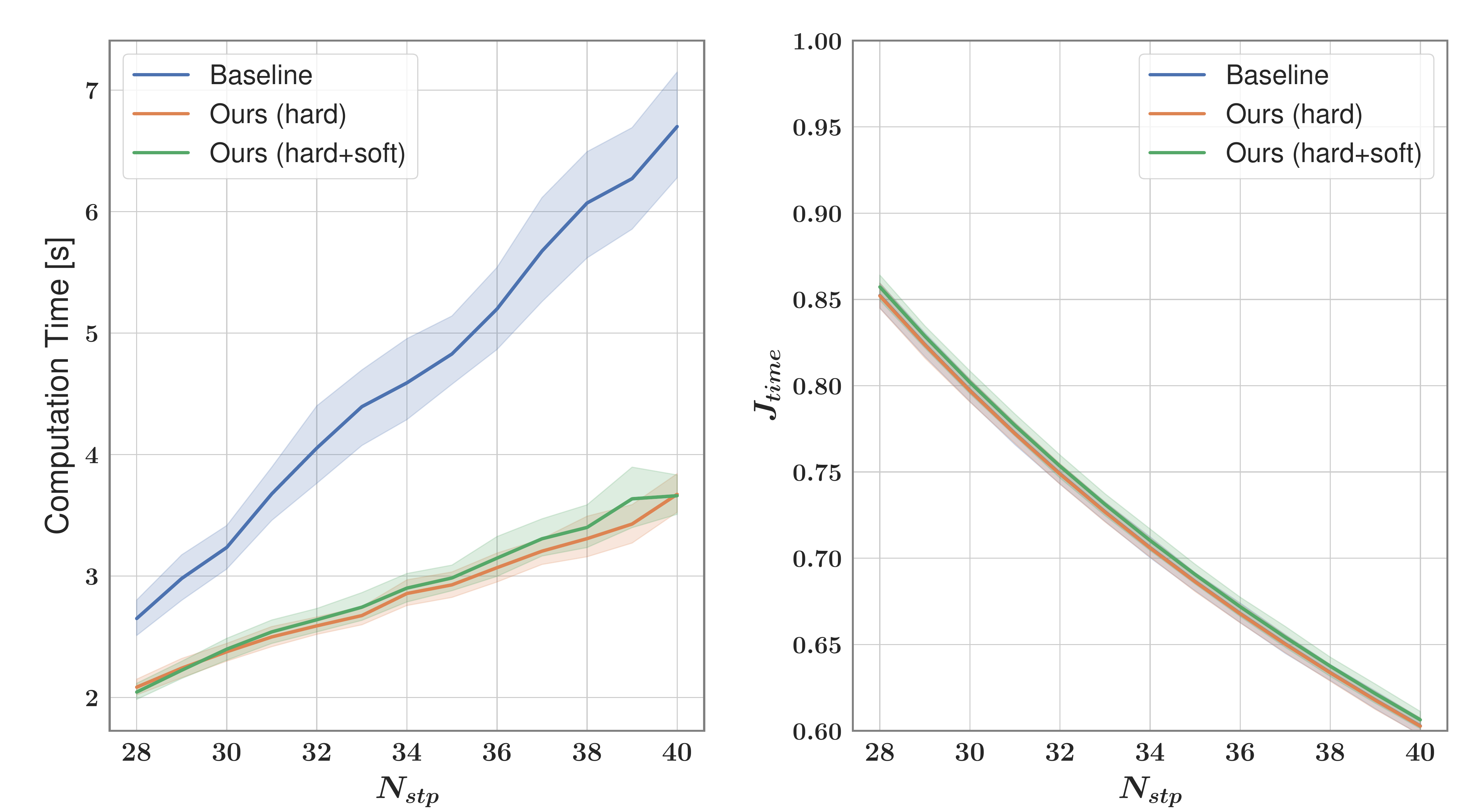}
    \vspace{-6mm}
    \caption{Sensitivity analysis on parameter $N_{stp}$ in the case of $N_{dlv}=2$. Each solid line represents the mean value. Each shaded area shows 99\% confidence intervals. Computation time means time to find an optimal solution.}
    \label{fig:result3}
\end{figure}

\begin{figure}[!t]
    \centering
    \includegraphics[keepaspectratio,clip,width=\hsize]{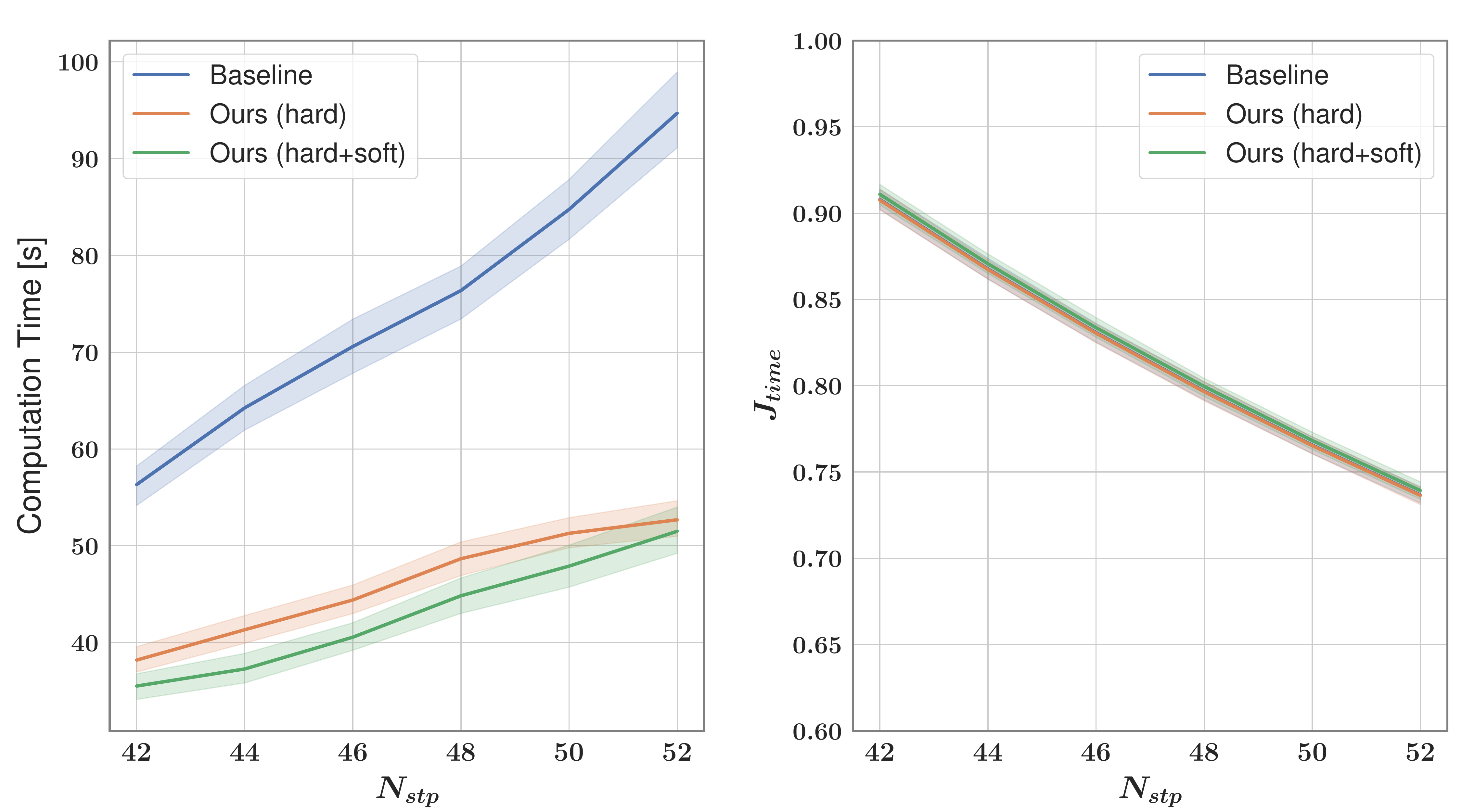}
    \vspace{-6mm}
    \caption{Sensitivity analysis on parameter $N_{stp}$ in the case of $N_{dlv}=3$. The graphic representation is in a similar manner with Fig. \ref{fig:result3}.}
    \label{fig:result4}
\end{figure}

Fig. \ref{fig:result3} and \ref{fig:result4} show the results of the sensitivity analysis on parameter $N_{stp}$ in the case of $N_{dlv}=2, 3$, respectively. We varied $N_{stp}$ from a tight setting ($J_{time}$ close to 1) to a loose setting (smaller $J_{time}$) for the use case where $N_{stp}$ cannot always be configured with a tight setting.
The left graph shows that the proposed models robustly outperform the baseline model in the range. Furthermore, in the case of $N_{dlv}=3$, ``Ours (hard+soft)'' is faster than ``Ours (hard)'' marginally but robustly.
The right graph shows that only ``Ours (hard+soft)'' resulted in slight degradation regarding the completion time steps $J_{time}$. This is the disadvantage of the second approach but its impact is acceptably small in practice.
We successfully validated effectiveness of the proposed models in the evaluation configuration which is a typical P\&P scenario.
However, further investigation with more complex configurations (\ie layout of obstacle, delivery, multi-arm robot, etc.) is needed. This is a future work to clarify how much the proposed model relies on a tradeoff between the completion time steps $J_{time}$ and the computation time.


%%%%%%%%%%%%%%%%%%%%%%%%%%%%%%%%%%%%%%%%%%%%%%%%%%%%%%%%%%%%%%%%%%%%%%%%%%%%%%%%

%%%%%%%%%%%%%%%%%%%%%%%%%%%%%%%%%%%%%%%%%%%%%%%%%%%%%%%%%%%%%%%%%%%%%%%%%%%%%%%%
%%%%%%%%%%%%%%%%%%%%%%%%%%%%%%%%%%%%%%%%%%%%%%%%%%%%%%%%%%%%%%%%%%%%%%%%%%%%%%%%
%%%%%%%%%%%%%%%%%%%%%%%%%%%%%%%%%%%%%%%%%%%%%%%%%%%%%%%%%%%%%%%%%%%%%%%%%%%%%%%%

\section{Conclusion}
We presented new models of MILP-based TAMP for robotic P\&P, which plans action sequences and motion trajectory with low computation costs. We improved an existing state-of-the-art MILP-based TAMP model integrated with collision avoidance. To enable the MILP solver to search for solutions efficiently, we introduced two approaches leveraging features of collision avoidance in robotic P\&P. The first approach reduces the number of binary variables, which are related to the collision avoidance of delivery objects, by reformulating them as continuous variables with additional hard constraints. These hard constraints maintain consistency by conditionally propagating binary values, which relate to the action state and collision avoidance of robots, to the reformulated continuous variables. The second approach is more aware of the branch-and-bound method which is the fundamental algorithm of modern MILP solvers. This approach is to guide the MILP solver to find integer solutions with shallower branching by adding a soft constraint, which softly restricts a robot’s routes around delivery objects. We demonstrated a considerable speed-up by the proposed models.

%%%%%%%%%%%%%%%%%%%%%%%%%%%%%%%%%%%%%%%%%%%%%%%%%%%%%%%%%%%%%%%%%%%%%%%%%%%%%%%%
\addtolength{\textheight}{-12cm}   % This command serves to balance the column lengths
                                  % on the last page of the document manually. It shortens
                                  % the textheight of the last page by a suitable amount.
                                  % This command does not take effect until the next page
                                  % so it should come on the page before the last. Make
                                  % sure that you do not shorten the textheight too much.

%%%%%%%%%%%%%%%%%%%%%%%%%%%%%%%%%%%%%%%%%%%%%%%%%%%%%%%%%%%%%%%%%%%%%%%%%%%%%%%%


\begin{thebibliography}{99}
%%%%%%%%%%%%%%%%%%%%%%%%%%%%%%%%%%%%%%
% Market-1
\bibitem{c01} International Federation of Robotics, 
 ``World robotics report 2020,''
 https://ifr.org/,
 2020.

%%%%%%%%%%%%%%%%%%%%%%%%%%%%%%%%%%%%%%
% TAMP:hierachical-1
\bibitem{c02} S. Srivastava, E. Fang, L. Riano, R. Chitnis, S. Russell, and P. Abbeel,
 ``Combined task and motion planning through an extensible planner-independent interface layer,''
 {\it IEEE Int. Conf. on Robotics and Automation (ICRA)},
 pp. 639--646, 2014.
% TAMP:hierachical-2
\bibitem{c03} T. L. Perez and P. Kaelbling,
 ``A constraint-based method for solving sequential manipulation planning problems,''
 {\it IEEE/RSJ Int. Conf. on Intelligent Robots and Systems (IROS)},
 pp. 3684--3691, 2014.
% TAMP:hierachical-3
\bibitem{c04} N. T. Dantam, Z. K. Kingston, S. Chaudhuri, and L. E. Kavraki,
 ``Incremental task and motion planning: a constraint-based approach,''
 {\it Robotics: Science and Systems (RSS)},
 2016.
% TAMP:hierachical-4
\bibitem{c05} C. Zhang and J. A. Shah,
 ``Co-optimizing task and motion planning,''
 {\it IEEE/RSJ Int. Conf. on Intelligent Robots and Systems (IROS)},
 pp. 4750--4756, 2016.

%%%%%%%%%%%%%%%%%%%%%%%%%%%%%%%%%%%%%%
% TAMP:optimization-1
\bibitem{c06} M. Toussaint,
 ``Logic-geometric programming: an optimization-based approach to combined task and motion planning,''
 {\it Int. Joint Conf. on Artificial Intelligence (IJCAI)},
 pp. 1930--1936, 2015.
% TAMP:optimization-2
\bibitem{c07} M. Toussaint and M. Lopes,
 ``Multi-bound tree search for logic-geometric programming in cooperative manipulation domains,''
 {\it IEEE Int. Conf. on Robotics and Automation (ICRA)},
  pp. 4044--4051, 2017.
% TAMP:optimization-3
\bibitem{c08} M. Katayama, S. Tokuda, M. Yamakita, and H. Oyama,
 ``Fast LTL-based flexible planning for dual-arm manipulation,''
 {\it IEEE/RSJ Int. Conf. on Intelligent Robots and Systems (IROS)},
 pp. 6605--6612, 2020.
% TAMP:optimization-4
\bibitem{c09} R. Takano, H. Oyama, and M. Yamakita,
 ``Continuous optimization-based task and motion planning with signal temporal logic specifications for sequential manipulation,''
 {\it IEEE Int. Conf. on Robotics and Automation (ICRA)},
 2021.

%%%%%%%%%%%%%%%%%%%%%%%%%%%%%%%%%%%%%%
% Collision:general-1
\bibitem{c10} X. Zhang, A. Liniger, and F. Borrelli,
``Optimization-based collision avoidance,''
{\it IEEE Transactions on Control Systems Technology},
 pp. 1--12, 2020.
 % Collision:free-1
\bibitem{c11} M. d. S. Atrantes, C. F. M. Toledo, B. C. Williams, and M. Ono,
``Collision-free encoding for chance-constrained nonconvex path planning,''
{\it IEEE Transactions on Robotics}, 
vol. 35, no. 2, pp. 433--448, 2019.
% Collision:free-2
\bibitem{c12} R. Deits and R. Tedrake,
 ``Efficient mixed-integer planning for UAVs in cluttered environments,''
 {\it IEEE Int. Conf. on Robotics and Automation (ICRA)},
 2015.
% Collision:free-3
\bibitem{c13} H. Dai, G. Izatt, and R. Tedrake,
 ``Global inverse kinematics via mixed-integer convex optimization,''
 {\it Int. Journal of Robotics Research,},
 vol. 38, no. 12--13, pp. 1420--1441, 2019.


%%%%%%%%%%%%%%%%%%%%%%%%%%%%%%%%%%%%%%
% MILP-history
\bibitem{c14} R. E. Bixby,
 ``A brief history of linear and mixed-integer programming computation,''
 {\it Documenta Mathematica: Optimization Stories},
 pp. 107-121, 2012.
 % MILP:performance-1
\bibitem{c15} Gurobi Optimization, LLC,
``Gurobi 9.0 performance benchmarks,''
https://www.gurobi.com/wp-content/uploads/2020/02/Performance-Gurobi-9.0-1.pdf,
2020.
% MILP:performance-2
\bibitem{c16} IBM Corporation,
``CPLEX optimization studio 12.10 performance improvements,''
https://www.ibm.com/downloads/cas/KEVDB4NZ,
2020.
% MILP-1
\bibitem{c17} J. P. Vielma,
 ``Mixed integer linear programming formulation techniques,''
 {\it SIAM review},
 Society for Industrial and Applied Mathematics, vol. 57, no. 1, pp. 3--57, 2015.
% MILP-1
\bibitem{c18} J. A. Huchette,
  ``Advanced mixed-integer programming formulations : methodology, computation, and application,''
  Ph.D. Dissertation, Massachusetts Institute of Technology, 2018.

%%%%%%%%%%%%%%%%%%%%%%%%%%%%%%%%%%%%%%
% Tool-1
\bibitem{c19} T. A. M. Toffolo and H. G. Santos,
 ``Python-mip,''
 GitHub repository, https://github.com/coin-or/python-mip,
 2021.
% Tool-2
\bibitem{c20} Gurobi Optimization, LLC,
 ``Gurobi optimizer reference manual,''
 https://www.gurobi.com/documentation/9.1/refman/index.html,
 2021.
%%%%%%%%%%%%%%%%%%%%%%%%%%%%%%%%%%%%%%
% Collision:etc-1

\end{thebibliography}
\end{document}